\documentclass[nohyperref]{article}

\usepackage{microtype}
\usepackage{graphicx}
\usepackage{subfigure}
\usepackage{booktabs} 

\usepackage{hyperref}

\usepackage{algorithm}
\usepackage{algorithmic}

\usepackage{comment}
\usepackage{enumitem}
\newcommand{\modelshort}{KCM}
\newcommand{\modelfull}{Kernelized Convex Masking}

\newcommand{\modelrank}{\textsl{R2D2}}
\newcommand{\RR}{\emph{R2}}
\newcommand{\RRfull}{Representative Ranking}
\newcommand{\DD}{\emph{D2}}
\newcommand{\DDfull}{Data-Driven}

\usepackage{amsmath}
\DeclareMathOperator*{\argmin}{argmin}

\usepackage[accepted]{icml2023}

\usepackage{amsmath}
\usepackage{amssymb}
\usepackage{mathtools}
\usepackage{amsthm}
\usepackage{multirow}
\usepackage{pifont}
\newcommand{\xmark}{\color{red} \ding{55}}
\definecolor{my-green}{rgb}{0.0, 0.5, 0.0}

\usepackage[capitalize,noabbrev]{cleveref}

\theoremstyle{plain}

\theoremstyle{definition}

\theoremstyle{remark}

\usepackage[textsize=tiny]{todonotes}

\icmltitlerunning{Gradient-Free Structured Pruning with Unlabeled Data}

\begin{document}

\twocolumn[
\icmltitle{Gradient-Free Structured Pruning with Unlabeled Data}
 


\icmlsetsymbol{equal}{*}

\begin{icmlauthorlist}
\icmlauthor{Azade Nova}{yyy}
\icmlauthor{Hanjun Dai}{yyy}
\icmlauthor{Dale Schuurmans}{yyy,comp}
\end{icmlauthorlist}

\icmlaffiliation{yyy}{Google DeepMind}
\icmlaffiliation{comp}{University of Alberta}

\icmlcorrespondingauthor{Azade Nova}{azade@google.com}

\icmlkeywords{Machine Learning, ICML}

\vskip 0.3in
]



\printAffiliationsAndNotice{} 

\begin{abstract}

Large Language Models (LLMs) have achieved great success in solving difficult tasks across many domains, but such success comes with a high computation cost, and inference latency. As developers and third parties customize these models, the need to provide efficient inference has increased. Many efforts have attempted to reduce inference cost through model compression techniques such as pruning and distillation. However, these techniques either require labeled data, or are time-consuming as they require the compressed model to be retrained to regain accuracy. In this paper, we propose a gradient-free structured pruning framework that uses only unlabeled data. An evaluation on the GLUE and SQuAD benchmarks using BERT$_{BASE}$ and DistilBERT illustrates the effectiveness of the proposed approach. By only using the weights of the pre-trained model and unlabeled data, in a matter of a few minutes on a single GPU, up to 40\% of the original FLOP count can be reduced with
less than a $4\%$ accuracy loss across all tasks considered. 

\end{abstract}

\section{Introduction}\label{sec:intro}
Large Language Models (LLMs) have made great strides in solving difficult tasks across many domains, but this has come at the cost of high parameter counts and significant computational overhead. Developers and third parties can now employ these trained models and create custom versions tailored to their particular applications. Customization makes these models applicable to a wider variety of use cases, but this, even more, highlights the need for efficient inference models.

Many efforts have been being made to reduce computational cost through model compression techniques specialized for Transformers, including structured pruning~\cite{xia2022structured, hou2020dynabert, sajjad2023effect, liu2021rosita, xia2022structured}, efficient architecture design~\cite{kitaev2020reformer, iandola2020squeezebert, sun2020mobilebert, wang2020linformer}, neural architecture search~\cite{so2021searching, xu2021bert, yin2021autotinybert}, knowledge distillation~\cite{sun2020mobilebert, jiao2019tinybert, sanh2019distilbert}, quantization~\cite{kim2021bert, shen2020q, zadeh2020gobo, zafrir2019q8bert}, and hardware-software co-design~\cite{gu2022heat, ham2021elsa}.

Among these techniques, structured pruning shows promising results in reducing model size, while also improving inference time because the resulting model remains compatible with the underlying hardware. However, most existing approaches are quite complex and require significant engineering effort to implement. Moreover, the process of compression is time-consuming and requires retraining the compressed model to regain accuracy. These limitations make effective compression difficult to realize in practice. Recently,~\citet{kwon2022fast} proposed a post-training pruning for Transformers that does not require any retraining of the model. Even though this approach avoids expensive retraining, it requires labeled data in the pruning pipeline. 

LLMs mainly utilize unlabeled data for training and with increased use of pre-trained LLMs by developers and third parties, access to the labeled data is questionable. Especially with the popularity of in-context learning, where the user only provides prompts, the purpose of the task is not necessarily known at compression time. In this scenario, none of the existing pruning techniques can be applied for model compression since they all require labeled data. Even though knowledge distillation~\cite{sun2020mobilebert, jiao2019tinybert, sanh2019distilbert} trains a student model with unlabeled data, it still requires a large amount of unlabeled data and is expensive to train. This motivates us to investigate whether one can design a structured pruning method that does not require retraining nor labeled data, while avoiding 
adverse 
effects
on performance. 

In this work, we propose \modelfull~(\modelshort), a gradient-free framework (Figure~\ref{fig:overview}) that only requires the trained model and sampled raw data to compress the model. We introduce
\modelrank~as the core of this framework that combines two ranking techniques \emph{\RRfull} (\RR) and \emph{\DDfull} (\emph{\DD}) to estimate the importance of individual neurons. \RR~maps our structured pruning goals into a representative selection problem~\cite{huang2018kernelized}, where the goal is to find a small subset of data points that can well represent a large dataset. Specifically, \RR~considers the filters of a Feed-Forward Network (FFN) in the trained model as data points in a high-dimensional space, and ranks these by how well a filter can be represented by others. \DD, on the other hand, ranks the filters based on statistics gathered from layer-wise model outputs using the raw sampled data. \modelshort~decides which filter to remove by merging the \modelrank~rankings across all layers. Since removing filters may still affect accuracy, we apply an existing scaling transformation method in~\citet{kwon2022fast} to 
mitigate the effect of their removal.

Our main contributions are as follows:
\begin{itemize}[noitemsep, leftmargin=*]
\item {We Propose \modelfull~(\modelshort)~ pruning framework, a gradient-free structured pruning approach that neither requires labeled data nor retraining.}
\item {As the core of \modelshort, we propose \modelrank~that combines two ranking techniques \emph{\RRfull} (\RR) and \emph{\DDfull} (\emph{\DD}). \modelrank~only requires the weights of the trained model and sampled raw data to rank the neurons. An ablation study confirms the importance of combining the two proposed ranking techniques.}
\item{Our evaluation on GLUE and SQuAD benchmarks using BERT$_{BASE}$ and DistilBERT confirms the effectiveness of the proposed approach. Compared to when the labeled data is available, \modelshort~is able to reduce up to 40\% of the original FLOPs with less than $4\%$ accuracy loss, in a matter of a few minutes on a single GPU.}

\end{itemize}

\section{Problem Definition}\label{sec:problem}
\subsection{Preliminary}
In this paper, we focus on pruning the BERT~\cite{devlin2018bert} architecture. BERT is a stack of $L$ homogeneous Transformer encoder blocks~\cite{vaswani2017attention}, each of which consists of a multi-head attention (MHA) layer followed by a Feed-Forward Network (FFN) layer. Due to the fact that FFN layers have a huge impact on model size and inference latency~\cite{ganesh2021compressing}, 
we focus on the pruning the filters of the FFN layers. In every transformer encoder layer $\ell$, the $FFN^{\ell}(x)$ with $N$ filters is parameterized with $W_{\ell}^{(1)} \in \mathbb{R}^{d \times N}$, $W_{\ell}^{(2)} \in \mathbb{R}^{N \times d}$, $b_{\ell}^{(1)}\in \mathbb{R}^{N}$, $b_{\ell}^{(2)}\in \mathbb{R}^{d}$, and activation function $\sigma$:
\begin{equation}
\label{equ:ffn}
\begin{aligned}
FFN_{\ell}(x) = \sum_{i=1} ^{N}(\sigma(xW^{(1)}_{\ell}[:,i] + b^{(1)}_{\ell})W^{(2)}_{\ell}[i,:]) + b^{(2)}_{\ell}
\end{aligned}
\end{equation}
For example, $BERT_{BASE}$ has 12 transformer encoder blocks ($L=12$), where the number of filters $(N)$ is 3072, $W_{\ell}^{(1)} \in \mathbb{R}^{768 \times 3072}$ with the GELU activation~\cite{hendrycks2016gaussian}, and $W_{\ell}^{(2)} \in \mathbb{R}^{3072 \times 768}$.

\begin{figure}
\centering
  \includegraphics[width=1.02\linewidth]{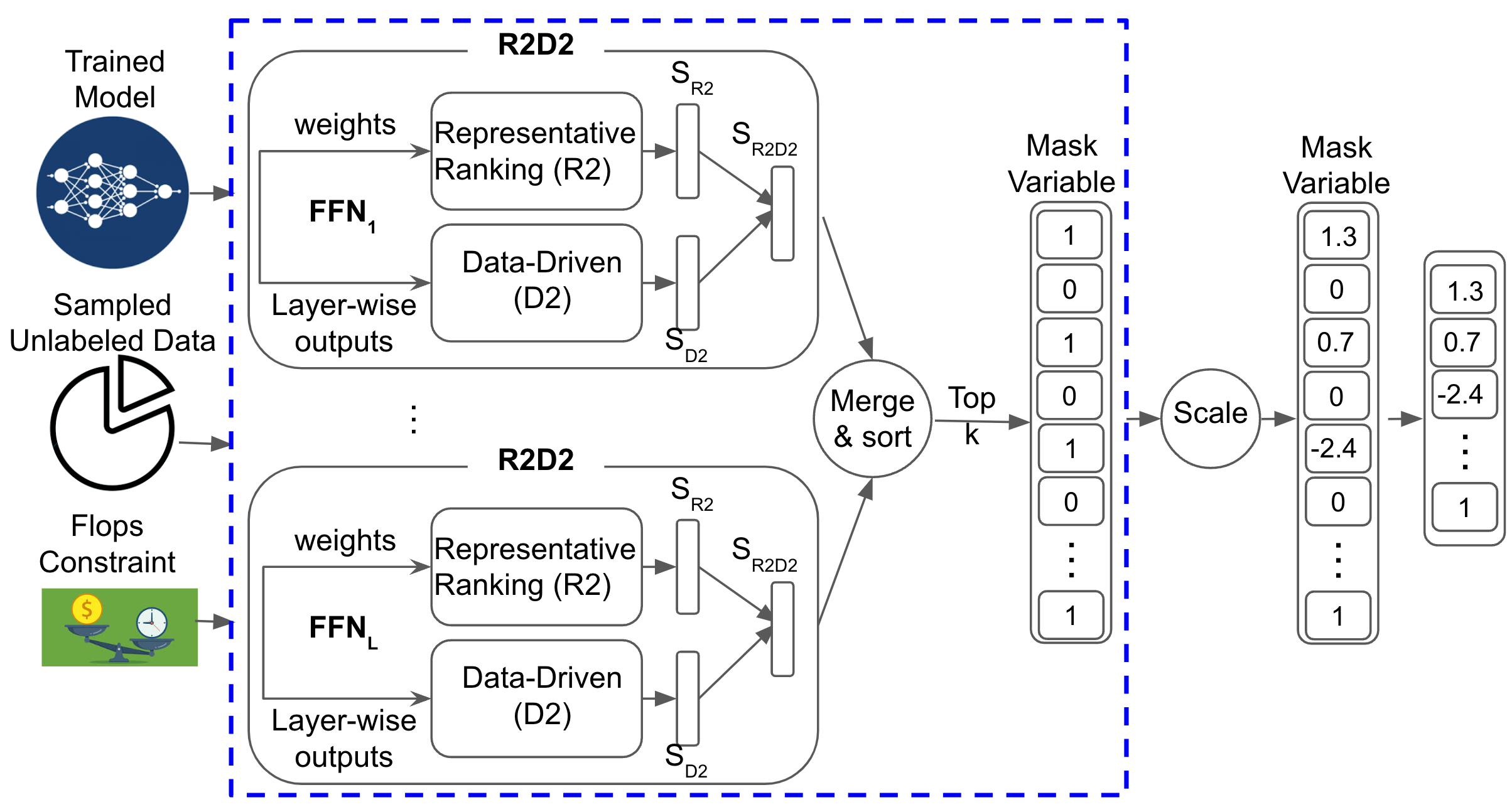}
  \vspace{-0.1in}
  \caption{\modelfull~(\modelshort): A gradient-free structured pruning framework with \modelrank~as a core component. \modelrank~combines the ranks of the \emph{\RRfull}~(\RR) and \emph{\DDfull}~(\DD) rank.}
  \label{fig:overview}
\end{figure}
\subsection{Structured Pruning by Masking}

\noindent{\bf Masking}: Given an integer $n < N$, reducing the number of filters from $N$ to $n$ can be considered as introducing a mask variable $m \in \mathbb{R}^{N}$ (with $n$ non-zero elements) associated with the outputs of the filters.
\begin{equation}
\label{equ:ffn-mask}
\begin{aligned}
\widehat{FFN}_{\ell}(x) = \sum_{i=1} ^{N}(\sigma(x W^{(1)}_{\ell}[:,i] + b^{(1)}_{\ell}) W^{(2)}_{\ell}[i,:] \circ m_i) + b^{(2)}_{\ell}
\end{aligned}
\end{equation}
where $\circ$ is Hadamard product.
%

\noindent{\bf Objective}: Transformer pruning can be formulated as a constrained optimization problem on the mask $\mathcal{M}\in \mathbb{R}^{L \times N}$ that represents the mask $m$ of all layers $L$. There are $LN$ filter mask variables which is much less than the total number of parameters in the model. For example BERT$_{BASE}$ with 110M parameters needs only 36k mask variables (0.03\%).

Optimal structural pruning is usually defined~\cite{kwon2022fast} in the supervised setting with respect to minimizing the accuracy loss of the original model: 
\begin{equation}
\label{equ:loss-supervised}
\begin{aligned}
\argmin_{\mathcal{M}} \mathcal{L}(\mathcal{M}) ~~~~s.t.~~~~ Cost(\mathcal{M}) \leq \mathcal{C}
\end{aligned}
\end{equation}
$Cost(\mathcal{M})$ is the floating point operations (FLOPs) of the pruned model determined by the mask $\mathcal{M}$. 
In this work, since only unlabeled data is available, the supervised loss $\mathcal{L}(\mathcal{M})$ can not be evaluated. 
Similar to distillation, we consider minimizing the Feature Map Loss~\cite{sun2020mobilebert} $\mathcal{L}_{FMT}$ for each FFN in layer $\ell$.
\begin{equation}
\label{equ:loss}
\begin{aligned}
\mathcal{L_{(\ell)}}_{FMT}(m) = \lVert FFN_{(\ell)}(x)- \widehat{FFN}_{(\ell)}(x)\rVert _2
\end{aligned}
\end{equation}
Given a trained model $\mathsf{Model}$, unlabeled dataset $\mathcal{D}$, and a cost constraint $\mathcal{C}$, find the mask $\mathcal{M} \in \mathbb{R}^{L \times N}$ for every $N$ filters of all $L$ transformer layers such that $Cost(\mathcal{M})$ be less than $\mathcal{C}$ and the loss $\mathcal{L}_{FMT}(\mathcal{M}) = \sum_{\ell=1}^{L} \mathcal{L^{(\ell)}}_{FMT}(\mathcal{M}_{\ell,:})$ is minimized.
\begin{equation}
\label{equ:opt}
\begin{aligned}
\argmin_{\mathcal{M}} \mathcal{L}_{FMT}(\mathcal{M}) ~~~~s.t.~~~~ Cost(\mathcal{M}) \leq \mathcal{C}
\end{aligned}
\end{equation}
One way to tackle this problem would be to consider it as a version of the distillation problem, where the goal is to find the optimal mask under the sparsity constraint. However, distillation methods require large amounts of unlabeled data and are very expensive to train~\cite{xia2022structured}.
\begin{algorithm}[t]
\small
\caption{\small \modelfull~(\modelshort)}
\begin{algorithmic} [1]
\label{alg:kcm}
\STATE{{\bf Input: } Trained model: $\mathsf{Model}$, FLOPs constraint $\mathcal{C}$, Gaussian Kernel $K$, convergence rate $\alpha$}
\STATE {{\bf Output: } Mask $\mathcal{M}$}
\STATE {Initialize mask $\mathcal{M}$ as $\mathbf{0}$}
\\ {\textit {//Call \RRfull~(\RR)~Algorithm~\ref{alg:frr}}}\\
\STATE{$S_{R2}$ = \RR($\mathsf{Model}$, $K$, $\alpha$)}
\\ {\textit {// \DDfull}~(\DD) Ranking}\\
\FOR{batch in sample-data}
\STATE{for each layer $\ell$ in $\mathsf{Model}$ collect $H_{\ell}^{(1)}$}
\STATE{$S_{D2}[\ell]$ = average over $H_{\ell}^{(1)}$ for each filter }
\ENDFOR
\STATE{$S_{R2D2}[\ell] = S_{R2}[\ell] * normalized(S_{D2}[\ell])$}
\STATE{$k$ = Number of neurons to satisfy FLOPs constraint $\mathcal{C}$}
\STATE{Candidates = top-$k$ filters of the sorted $S_{R2D2}$}
\STATE{$\mathcal{M}[Candidates] = 1.0$}
\STATE{{\bf return } $\mathcal{M}$}
\end{algorithmic}
\end{algorithm}
\begin{algorithm}[t]
\small
\caption{\small \RRfull~(R2) }
\begin{algorithmic} [1]
\label{alg:frr}
\STATE{{\bf Input: } Trained model: $\mathsf{Model}$, Gaussian Kernel $K$ with width $\sigma$, convergence rate $\alpha$}
\STATE {{\bf Output: } $S_{R2}$ that represents importance of Filters in all layers.}
\FOR{ layer $\ell$ in layers of the $\mathsf{Model}$}
\STATE {$W_{\ell}^{(2)} \in \mathbb{R}^{N \times d}$ of $FFN_{\ell}$}%
\STATE {Initialize coefficient matrix: $C_0 \in \mathbb{R}^{N \times N} = \frac {1}{N}$}
\REPEAT
    \STATE{$C_{i+1} = C_{i} \circ \sqrt{\frac
                                       {K(W_{\ell}^{(2)}, W_{\ell}^{(2)})}
                                       {K(W_{\ell}^{(2)}, W_{\ell}^{(2)})C_i}}$}
    \STATE{$\delta = \frac {(C_{i+1} - C_{i}).sum()} {C_{i}.sum}$}
    \STATE{$C_{i} = C_{i+1}$}
\UNTIL{convergence i.e.  $\delta \leq \alpha$ }
\STATE{$S_{R2}[{\ell}]$= diagonal($C_i$)}
\ENDFOR
 \STATE{{\bf return } $S_{R2}$}
\end{algorithmic}
\end{algorithm}
\section{Proposed Approach}\label{sec:approach}
Instead, in this work, we propose a gradient-free approach that only uses the weights of the trained model and statistics on layer-wise outputs using the unlabeled data to implicitly minimizes the feature map loss in each layer.
\subsection{Framework Overview}
Figure~\ref{fig:overview} shows the overview of our framework, called \modelfull~(\modelshort). \modelshort~takes the trained model $\mathsf{Model}$, sampled unlabeled dataset $\mathcal{D}$ and a cost constraint $\mathcal{C}$, and returns a mask $\mathcal{M}\in \mathbb{R}^{L \times N}$ that represents the mask of the $N$ filters of all layers $L$. 

 We introduce \modelrank~that combines ranking techniques \emph{\RRfull} (\RR) and \emph{\DDfull} (\emph{\DD}) to estimate the importance of the filters. As shown in Figure~\ref{fig:overview}, these two approaches independently rank $N$ filters based on the weights and output of the activation function of the FFNs in all layers $L$. Then \modelshort~merges the results of \modelrank~across all layers. The top $k$ filters are 
 selected and the rest will be masked to zero. Note that given a FLOPs constraint $\mathcal{C}$, $k$ is the total number of filters that satisfies constraint $\mathcal{C}$. Finally, we apply a scaling transformation in~\citet{kwon2022fast} over the selected filters to recover the accuracy drop and reduce the feature map loss in Equation~\ref{equ:loss}. Next, we discuss our framework in more detail. 

\subsection{\modelfull (\modelshort)}
Algorithm~\ref{alg:kcm} illustrates the end-to-end approach. We first present the details of the proposed \modelrank. Then, we discuss how we use these rankings for the final masking.    
\begin{table*}[h!]
    \centering
    \caption{Comparison of the different structured pruning methods studied in this work. \xmark \color{black}, and \color{my-green} \checkmark \color{black} show if a method has the specific feature or not. N/A means not applicable. To simplify notation, we show gradient-free with ($!\nabla$). Supervision-free indicates not using labeled data.}
    \label{tab:baselines}
\resizebox{0.85\textwidth}{!}{%
    \begin{tabular}{ccccc}
    \toprule
    Method & Gradient-free ($!\nabla$) & Retrain/Finetune-free & Supervision-free & Pruning time $\leq 7min$\\
    \hline
    FLOP~\cite{wang2019structured} & \xmark & \xmark & \xmark & \xmark\\
    SLIP~\cite{lin2020pruning} & \xmark & \xmark & \xmark & \xmark\\
    Sajjad et al.~\cite{sajjad2023effect} & \xmark & \xmark & \xmark & \xmark\\
    DynaBERT~\cite{hou2020dynabert} & \xmark & \xmark & \xmark & \xmark\\
    EBERT~\cite{liu2021ebert} & \xmark & \xmark & \xmark & \xmark\\
    \hline
    Mask-Tuning~\cite{kwon2022fast} & \xmark & \color{my-green} \checkmark & \xmark & \color{my-green} \checkmark\\
    \hline
    Weight-Magnitude~\cite{li2016pruning} & \color{my-green} \checkmark & \color{my-green} \checkmark & N/A & \color{my-green} \checkmark\\
    Weight-Magnitude-Scale  & \color{my-green} \checkmark & \color{my-green} \checkmark & \color{my-green} \checkmark & \color{my-green} \checkmark\\
    \bf \modelshort~(ours) & \color{my-green} \checkmark & \color{my-green} \checkmark & \color{my-green} \checkmark & \color{my-green} \checkmark\\
    \bottomrule
    \end{tabular}%
}
\end{table*}
\subsubsection{\RRfull~(\RR)}
By considering $H_{\ell}^{(1)} = \sigma(xW^{(1)}_{\ell} + b_{\ell}^{(1)})$, Equation~\ref{equ:ffn} can be written as $FFN_{\ell}(H^{(1)}_{\ell}) = H_{\ell}^{(1)}W_{\ell}^{(2)} + b_{\ell}^{(2)}$. Our filter \emph{\RRfull} assumes $H_{\ell}^{(1)}$ is unknown and only uses the weights $W_{\ell}^{(2)}$ to rank the $N$ filters.

From the computational geometry perspective the filters in $W_{\ell}^{(2)} \in \mathbb{R}^{N \times d}$ can be considered as $N$ data points in a $d$ dimensional space. The structured pruning goal can be translated as selecting a subset of data points (filters) to be used as representatives that can describe any data point (filter) in the dataset. There has been a lot of work on finding such a representative set~\cite{kazemi2022tackling, killamsetty2021retrieve, you2020self}. However, for linear functions, this problem can be reduced to finding a convex hull. \emph{The convex hull is a subset of data points that can be used to find the maxima of any linear function.} Since $FFN_{\ell}(H^{(1)}_{\ell})$ is, in fact, a linear function, the convex hull of $W_{\ell}^{(2)}$ can be considered as a representative of the filters that produce the maxima of $FFN_{\ell}$ regardless of the input $H^{(1)}_{\ell}$. 

The challenge is that in a $d$ dimensional space finding the exact convex hull is in order of $\mathcal{O}(N^{d/2})$ time, which can be very expensive (e.g. in $BERT_{BASE}$, d=768). Moreover, the number of convex hull data points radically increases with the number of dimensions. To address these limitations, rather than finding the exact solution, we propose to assign a ranking over the filters that represents how well a filter is representing others. 

Convex hull approximation is well-studied area. Among existing methods \emph{Kernelized Convex Hull Approximation (KCHA)}~\cite{huang2018kernelized} is one of the approaches that can be applied to our problem. Algorithm~\ref{alg:frr} shows the proposed \emph{\RRfull} based on the KCHA. 
Specifically for each layer $\ell$, we seek a positive coefficient matrix $C \in \mathbb{R}^{N \times N}$ that minimizes $\lVert W_{\ell}^{(2)} - W_{\ell}^{(2)}C\rVert_2$. The diagonal elements of $C$ indicate whether the corresponding data instances are extreme points. 
\citet{huang2018kernelized} solves this problem as a Semi-NMF problem~\cite{ding2008convex}, rather than a Non-negative Least Square problem, and adopts a multiplicative updating rule as the solver:
\begin{equation}
\label{equ:updating-rule}
\begin{aligned}
C^{i+1} = C^{i} \circ \sqrt{
\frac
{[W_{\ell}^{(2)^T}W_{\ell}^{(2)}]_{+} + [W_{\ell}^{(2)^T}W_{\ell}^{(2)}]_{-}C^i}
{[W_{\ell}^{(2)^T}W_{\ell}^{(2)}]_{-} + [W_{\ell}^{(2)^T}W_{\ell}^{(2)}]_{+}C^i}
                            }
\end{aligned}
\end{equation}
where $[A]_+=\frac{A+|A|}{2}$, $[A]_-=\frac{A-|A|}{2}$, and $|A|$ is the absolute values of $A$. Please refer to~\citet{huang2018kernelized} for more detail.

$W_{\ell}^{(2)^T}W_{\ell}^{(2)}$ can be considered as $K(W_{\ell}^{(2)}, W_{\ell}^{(2)})$. In this paper, we use a Gaussian kernel. Since the kernel value is positive and $K(W_{\ell}^{(2)}, W_{\ell}^{(2)}) = 1$, the updating rule of Semi-NMF algorithm can be modified as:
\begin{equation}
\label{equ:kernel-updating-rule}
\begin{aligned}
C^{i+1} = C^{i} \circ \sqrt{
\frac
{K(W_{\ell}^{(2)}, W_{\ell}^{(2)})}
{K(W_{\ell}^{(2)}, W_{\ell}^{(2)})C^i}
                           }
\end{aligned}
\end{equation}
Algorithm~\ref{alg:frr} illustrates the steps of the \emph{\RRfull}, where for each layer the coefficient matrix $C$ is independently calculated using Equation~\ref{equ:kernel-updating-rule}. The algorithm then returns the diagonal of $C$ as the ranking score of the filters. The width of the Gaussian kernel $\sigma$, and the convergence rate $\alpha$ are hyperparameters. In our experiments, we observe that setting $\sigma=1.0$ and $\alpha=0.01$ works for all tasks considered. Moreover, on average it takes less than 20 iterations to converge.   
\subsubsection{\DDfull~Ranking (\DD)}
\RRfull~(\RR) assumes $H_{\ell}^{(1)}$ is unknown and ranks the filters solely based on the weights $W_{\ell}^{(2)}$. One could imagine using a similar ranking approach over $W_{\ell}^{(1)^T}$ to rank the $N$ filters. 
However, as mentioned, the convex hull is only a good representative for finding the maxima of any \emph{linear function}, and the activation function $\sigma$ makes $H_{\ell}^{(1)}$ nonlinear.

Therefore, to incorporate the nonlinearity introduced by the activation function, \emph{\DDfull} (\DD) performs a forward pass using sampled unlabeled data, and gathers statistics on the output results of each layer $H_{\ell}^{(1)}$. It then uses the normalized average of these outputs to rank filters in each layer (Algorithm~\ref{alg:kcm} lines (5 to 7)).
\subsubsection{Merge and Scale}
Thus far, \modelshort~ranks $N$ filters of each layer independently by the filter \RRfull~(\RR) and \DDfull\  Ranking~(\DD). In every layer, \modelrank~combines the scores of \RR~and \DD~to capture the importance of filters based on the model weights and the layer outputs of the raw data (Algorithm~\ref{alg:kcm} line 9). In our experiments, we run an ablation study to present the importance of these rankings.

Given a FLOPs constraint $\mathcal{C}$, let $k$ be the total number of filters that satisfy $\mathcal{C}$. In other words, the pruned model only should have $k$ active filters across all layers and the rest should be removed. As shown in Algorithm~\ref{alg:kcm}, \modelshort~merges the \modelrank~scores ($S_{R2D2}$) across layers, and the top $k$ filters are selected to be active in the pruned model (Algorithm~\ref{alg:kcm} lines 10-12). 

Since after masking some accuracy drop is inevitable, existing structured pruning methods have shown that scaling can be helpful. Thus, similar to~\citet{kwon2022fast} as shown in Figure~\ref{fig:overview}, we apply a scaling transformation to the selected filters. Such scaling uses only the unlabeled data and, based on the generated mask, aims to reconstruct the layer-wise outputs by scaling the outputs of the active filters. This, in fact, reduces the feature map loss in Equation~\ref{equ:loss}.


\begin{table*}[h!]
    \centering
    \caption{Accuracy degradation of pruning BERT$_{BASE}$ using our method and the prior structured pruning methods with different relative FLOPs. Note that our method is gradient-free ($!\nabla$), does not use label of the data and not require retraining (more detail in Table~\ref{tab:baselines})}
    \label{tab:accuracy-drop}
\resizebox{1.0\textwidth}{!}{%
    \begin{tabular}{c|c|ccc|ccc|ccc|ccc}
    \toprule
    \multirow{2}{*}{$!\nabla$} & \multirow{2}{*}{Method} &   \multicolumn{3}{c|}{QQP} & \multicolumn{3}{c|}{QNLI} & \multicolumn{3}{c|}{SST-2} & \multicolumn{3}{c}{MRPC} \\
    & &  $\sim$60\% & $\sim$65\% & 75\% & $\sim$60\% & $\sim$65\% & 75\% & $\sim$60\% & $\sim$65\% & 75\% & $\sim$60\% & $\sim$65\% & 75\% \\
    \hline
    \xmark & FLOP &  $-$ & $-$ & $-$ & $-$ & -2.6 & $-$ & $-$ & -0.6 & $-$  & $-$ & -2.3 & $-$\\
    \xmark & SLIP &  -1.7 & -0.9 & $-$ & -2.1 & -0.9 & $-$ & -1.0 & -0.9 & $-$  & -1.0 & -2.8 & $-$\\
    \xmark & Sajjad &  $-$ & -0.4 & $-$ & $-$ & -1.4 & $-$ & $-$ & -1.8 & $-$  & $-$ & -8.6 & $-$\\
    \xmark & DynaBERT &  $-$ & $-$ & $-$ & $-$ & $-$ & $-$ & $-$ & $-$ & -0.6  & $-$ & $-$ & -1.7\\
    \xmark & EBERT &  -0.4 & $-$ & $-$ & -1.3 & $-$ & -1.0 & $-$ & $-$ & $-$  & $-$ & $-$ & $-$\\
    \hline
    \xmark & Mask-Tuning &  -0.65 & -0.42 & -0.14 & -1.35 & -0.89 & -0.18 & -1.08 & -0.91 & -0.68  & -1.72 & -0.24 & -0.14\\
    \hline
    {\color{my-green} \checkmark} & \bf \modelshort~(Ours) & -1.85 & -.99 & -0.35 & -3.62 & -1.72 & -0.78 & -2.46 & -1.71 & -0.68  & -1.96 & -1.47 & -0.71\\
    \bottomrule
    \end{tabular}%
}
\end{table*}
\begin{figure*}[h!]
\centering
  \includegraphics[width=\linewidth]{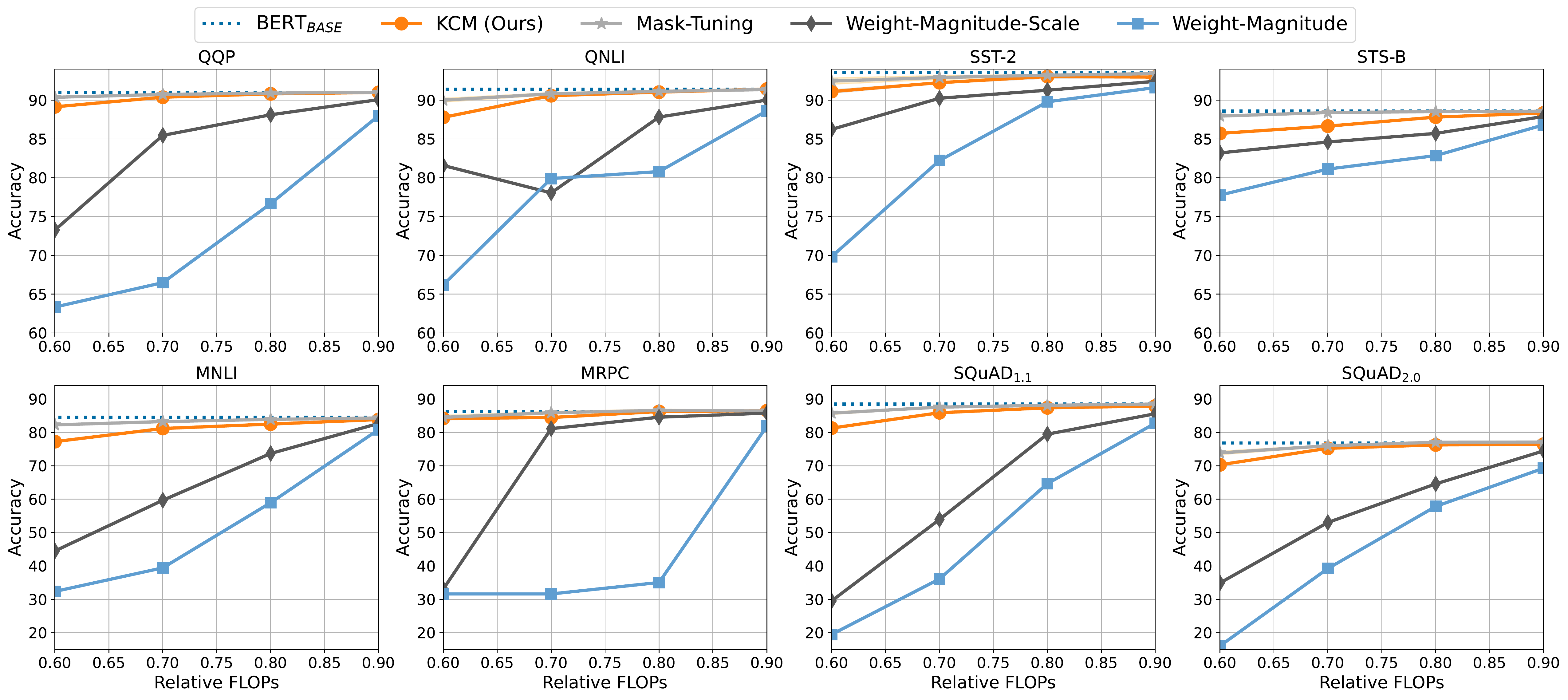}
  \caption{Performance of our pruning framework \modelshort~against Mask-Tuning~\cite{kwon2022fast}, Weight-Magnitude~\cite{li2016pruning}, and Weight-Magnitude-Scale on BERT$_{BASE}$. Weight-Magnitude-Scale combines~\cite{li2016pruning} with the scaling approach from~\cite{li2016pruning}. Mask-Tuning uses labeled data but \modelshort~and Weight-Magnitude-Scale are gradient-free with unlabeled data (Table~\ref{tab:baselines}). \modelshort~outperforms Weight-Magnitude and Weight-Magnitude-Scale which highlights the effectiveness of our approach in the absence of labeled data. For 70\% and 60\% FLOPs constraints, Mask-Tuning that uses labeled data performs slightly better than \modelshort. Table~\ref{tab:bert_1} shows this gap more clearly.}
  \label{fig:bert_res2}
\end{figure*}
\section{Evaluation}\label{sec:eval}
\subsection{Experimental Setup}
We implemented our framework with PyTorch~\cite{pytorch} using the HuggingFace Transformers~\cite{hug_transformer} library. We evaluate the effectiveness of the proposed approach using BERT$_{BASE}$~\cite{devlin2018bert} and DistilBERT~\cite{sanh2019distilbert} on GLUE~\cite{wang2018glue} and SQuAD~\cite{rajpurkar2018know, rajpurkar2016squad} benchmarks. 

For the \DDfull~ranking we use 2K raw data from the training sets. Note that we only use raw input and no label is used. Figure~\ref{fig:sample_num}, in Appendix~\ref{lbl:exp-results-appendix}, shows how sample size affects our performance. For the \RRfull~calculation in Algorithm~\ref{alg:frr}, the width of the Gaussian kernel, $\sigma$, and the convergence rate, $\alpha$, are the hyperparameters. In our experiments, we set $\sigma=1.0$ and $\alpha=0.01$. Moreover, on average it takes less than 20 iterations to converge. All results are averaged over the runs with 10 different seeds. Please refer to Appendix~\ref{lbl:exp-detail-appendix} for more detail on experimental setup.
\begin{table*}[h!]
    \centering
    \caption{Performance of our pruning framework \modelshort~against Mask-Tuning~\cite{kwon2022fast} on BERT$_{BASE}$, for 70\% and 60\% FLOPs constraints. Mask-Tuning that uses labeled data performs slightly better than \modelshort. Unlike Mask-Tuning, \modelshort~is gradient-free($!\nabla$) with unlabeled data. }
    \label{tab:bert_1}
\resizebox{1.0\textwidth}{!}{%
    \begin{tabular}{c|c|cc|cc|cc|cc}
    \toprule
    \multirow{2}{*}{$!\nabla$} & \multirow{2}{*}{Method} &  \multicolumn{2}{c|}{QQP} & \multicolumn{2}{c|}{MNLI} & \multicolumn{2}{c|}{MRPC} & \multicolumn{2}{c}{QNLI} \\
    &  & 60\% & 70\%  & 60\% & 70\% & 60\% & 70\% & 60\%  & 70\% \\
    \hline
    & baseline & \multicolumn{2}{c|} {91.00} & \multicolumn{2}{c|} {84.53} & \multicolumn{2}{c|} {86.27} & \multicolumn{2}{c} {91.41} \\
    \hline
    \xmark & Mask-Tuning   & $90.38 \pm 0.07$ & $90.74 \pm 0.07$ & $82.26 \pm 0.21$ & $83.24 \pm 0.16$ & $84.51 \pm 0.63$ & $85.91 \pm 0.40$ & $90.00 \pm 0.26$ & $90.83 \pm 0.16$\\
    \color{my-green} \checkmark & \bf \modelshort~(Ours) &  $89.15 \pm 0.04$ & $90.39 \pm 0.04$ & $77.24 \pm 0.10$ & $81.18 \pm 0.10$ & $84.19 \pm 0.44$ & $84.46 \pm 0.29$ & $87.79 \pm 0.15$ & $90.58 \pm 0.08$\\
    \hline
    \hline
    \multirow{2}{*}{$!\nabla$} &\multirow{2}{*}{Method} &  \multicolumn{2}{c|}{SST-2} & \multicolumn{2}{c|}{STS-B} & \multicolumn{2}{c|}{SQuAD$_{1.1}$} & \multicolumn{2}{c}{SQuAD$_{2.0}$} \\
    &  & 60\% & 70\%  & 60\% & 70\% & 60\% & 70\% & 60\%  & 70\% \\
    & baseline & \multicolumn{2}{c|} {93.57} & \multicolumn{2}{c|} {88.59} & \multicolumn{2}{c|} {88.48} & \multicolumn{2}{c} {76.82} \\
    \hline
    \xmark & Mask-Tuning   & $92.47 \pm 0.41$ & $92.92 \pm 0.26$ & $87.95 \pm 0.12$ & $88.40 \pm 0.05$ & $85.77 \pm 0.41$ & $87.57 \pm 0.11$ & $73.86 \pm 0.55$ & $76.00 \pm 0.29$\\
    \color{my-green} \checkmark & \bf \modelshort~(Ours) &  $91.11 \pm 0.23$ & $92.26 \pm 0.09$ & $85.72 \pm 0.12$ & $86.66 \pm 0.05$ & $81.29 \pm 0.06$ & $85.89 \pm 0.04$ & $70.30 \pm 0.13$ & $75.24 \pm 0.10$\\
    \bottomrule
    \end{tabular}%
}
\end{table*}
\begin{table*}[h!]
    \centering
    \caption{Performance of our pruning framework \modelshort~against Mask-Tuning~\cite{kwon2022fast} on DistilBERT, for 70\% and 60\% FLOPs constraints. Even though Mask-Tuning uses labeled data, for SQUAD$_{2.0}$ task, \modelshort~performs better than Mask-Tuning, and the results of both approaches on QQP, and STS-B are comparable.}
    \label{tab:distill-bert_1}
\resizebox{1.0\textwidth}{!}{%
    \begin{tabular}{c|c|cc|cc|cc|cc}
    \toprule
    \multirow{2}{*}{$!\nabla$} & \multirow{2}{*}{Method} &  \multicolumn{2}{c|}{QQP} & \multicolumn{2}{c|}{MNLI} & \multicolumn{2}{c|}{MRPC} & \multicolumn{2}{c}{QNLI} \\
    &  & 60\% & 70\%  & 60\% & 70\% & 60\% & 70\% & 60\%  & 70\% \\
    \hline
    & baseline & \multicolumn{2}{c|} {89.99} & \multicolumn{2}{c|} {82.11} & \multicolumn{2}{c|} {84.80} & \multicolumn{2}{c} {88.56} \\
    \hline
    \xmark & Mask-Tuning & $88.71 \pm 0.22$ & $89.66 \pm 0.06$ & $80.51 \pm 0.19$ & $81.65 \pm 0.09$ & $84.73 \pm 0.71$ & $84.83 \pm 0.35$ & $87.72 \pm 0.38$ & $88.43 \pm 0.07$  \\
    \color{my-green} \checkmark & \bf \modelshort~(Ours) & $88.16 \pm 0.03$ & $89.28 \pm 0.03$ & $78.05 \pm 0.08$ & $80.60 \pm 0.05$ & $79.66 \pm 0.27$ & $83.01 \pm 0.16$ & $85.93 \pm 0.09$ & $86.93 \pm 0.13$ \\
    \hline
    \hline
    \multirow{2}{*}{$!\nabla$} &\multirow{2}{*}{Method} &  \multicolumn{2}{c|}{SST-2} & \multicolumn{2}{c|}{STS-B} & \multicolumn{2}{c|}{SQuAD$_{1.1}$} & \multicolumn{2}{c}{SQuAD$_{2.0}$} \\
    & &  60\% & 70\%  & 60\% & 70\% & 60\% & 70\% & 60\%  & 70\% \\
    \hline
    & baseline & \multicolumn{2}{c|} {91.40} & \multicolumn{2}{c|} {86.12}  & \multicolumn{2}{c|} {85.73} & \multicolumn{2}{c} {68.84}\\
    \hline
    \xmark & Mask-Tuning & $90.44 \pm 0.41$ & $90.93 \pm 0.24$ & $85.73 \pm 0.07$ & $85.96 \pm 0.10$ & $83.20 \pm 0.16$ & $84.64 \pm 0.09$ & $62.36 \pm 1.40$ & $65.32 \pm 0.48$\\
    \color{my-green} \checkmark & \bf \modelshort~(Ours) & $88.38 \pm 0.25$ & $90.61 \pm 0.25$ & $85.26 \pm 0.02$ & $85.55 \pm 0.03$ & $76.92 \pm 0.11$ & $82.65 \pm 0.06$ & \bf{64.56 $\pm$ 0.11} & \bf{68.19 $\pm$ 0.06}\\
    \bottomrule
    \end{tabular}%
}
\end{table*}

\noindent{\bf Baselines from structured pruning methods}: 
Table~\ref{tab:baselines}, shows the comparison of the different structured pruning methods specialized for Transformers studied in this work. We compare these methods by 4 important features, including gradient-free (no backward pass), retrain/finetune-free (no retrain/finetune), supervision-free (no use of labeled data), and fast pruning-time.
We compare our proposed method with Flop~\cite{wang2019structured}, SLIP~\cite{lin2020pruning}, Sajjad et al.~\cite{sajjad2023effect}, DynaBERT~\cite{hou2020dynabert}, and EBERT~\cite{liu2021ebert}. All of these techniques require retraining of the pruned model and/or jointly learning the pruning configurations during training, which leads to high training time, and they are not gradient-free. Specifically, as shown in~\citet{kwon2022fast} these methods require 5 to 33 hours of retraining.

Mask-Tuning~\cite{kwon2022fast} is a recent work that does not need retraining but still relies on the labeled data and uses gradient computation to evaluate the importance of each filter. We also compare our method with Weight-Magnitude~\cite{li2016pruning}, which is a light-structured pruning method that is gradient-free, does not retrain the pruned model, and does not use the data at all. We introduce Weight-Magnitude-scale that combines~\citet{li2016pruning} with the scaling approach from~\citet{li2016pruning}. Note that the scaling step in~\citet{li2016pruning} only needs unlabeled data, so Weight-Magnitude-scale will have the exact problem setup as our method~\ref{tab:baselines}. We would like to highlight that our method \modelshort, Mask-Tuning, Weight-Magnitude, and Weight-Magnitude-scale finish in less than 7 minutes across all tasks, which is 2 to 3 orders of magnitude faster than the other baselines.
We evaluate the performance of our method against all these baselines by the FLOPs-accuracy trade-off of BERT$_{BASE}$ on the GLUE and SQuAD benchmarks. In the experimental results, to simplify the notation, we will indicate gradient-free with ($!\nabla$).
\begin{figure*}[t]
\centering
  \includegraphics[width=\linewidth]{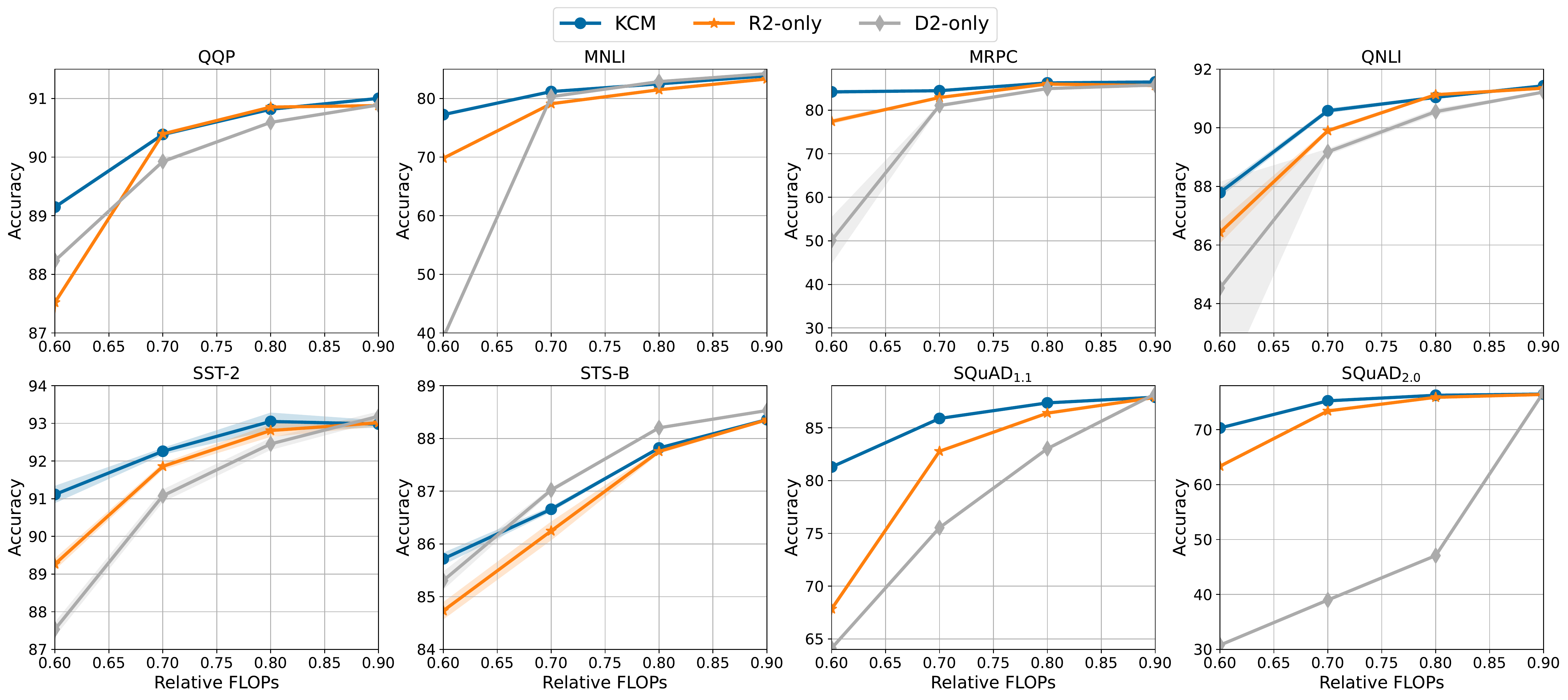}
  \caption{Ablation Study to investigate the importance of our ranking techniques. D2-only, and R2-only are our \modelshort~that either uses \emph{\DDfull} or \emph{\RRfull}. R2-only performs better than D2-only in all tasks except STS-B. But \modelshort~that combines them by \modelrank~is able to leverage both of these rankings and shows improvement across all tasks considered.}
  \label{fig:ablation}
\end{figure*}
\subsection{Experimental Results}
Table~\ref{tab:accuracy-drop} compares the accuracy drop of \modelshort~against prior structured
pruning methods in Table~\ref{tab:baselines}. Since the baseline accuracy differs slightly from paper to paper, we compare the amount of the accuracy drop from the baseline instead of the absolute accuracy. Similar to~\citet{kwon2022fast}, we use the results without knowledge distillation and data augmentation reported in each paper since these add extra overhead. As one can see, the highest accuracy drop of \modelshort~across all task is $-3.62$ which reduces 40\% of the original FLOPs. Worth mentioning that, while all the baselines require labeled data and leverage the backward pass, our proposed method is gradient-free with unlabeled data.

Next, we perform a more thorough evaluation against Mask-Tuning~\cite{kwon2022fast}, Weight-Magnitude~\cite{li2016pruning} and Weight-Magnitude-Scale, since their problem setup is closer to ours (Table ~\ref{tab:baselines}). Figure~\ref{fig:bert_res2} shows the results on BERT$_{BASE}$ as we vary the FLOPs constraint from 90\% to 60\%, i.e, reducing 10\% to 40\% of the original FLOPs. Clearly, \modelshort~outperforms Weight-Magnitude and Weight-Magnitude-Scale, highlighting the effectiveness of our approach in the absence of labeled data. For the 70\% and 60\% FLOPs constraints, Mask-Tuning performs slightly better than ours. Table~\ref{tab:bert_1} shows the gap more clearly. This gap can be explained by the fact that, unlike Mask-Tuning, \modelshort~is gradient-free with unlabeled data.

We further evaluate the performance of \modelshort~against Mask-Tuning~\cite{kwon2022fast} on DistilBERT for the 70\% and 60\% FLOPs constraints. As shown in Table~\ref{tab:distill-bert_1}, interestingly, even though Mask-Tuning leverages the backward pass and labeled data, the proposed \modelshort~performs better than Mask-Tuning on the SQUAD$_{2.0}$ benchmark. Moreover the results of both approaches on QQP, and STS-B are quite comparable, showing that even without labeled data and no backward pass the accuracy loss of the pruned model by our \modelshort~method~can be minimal.

\subsection{Ablation Studies}
\noindent{\bf Importance of our ranking techniques}:
\modelrank~is the core component of our proposed approach \modelshort~that combines the ranking of \RRfull~(\RR) and \DDfull~(\DD) (Section~\ref{sec:approach}). We run an ablation study to investigate the importance of these ranking techniques.
Recall that \RR~ranks $N$ filters based on the weights of the FFNs, while \DD~ranks them by the output of the activation function. Figure~\ref{fig:ablation} illustrates how the performance changes if we only use one of these in our framework. While \DD-only is our \modelshort~without the \RR, \RR-only only uses the \RR~ranking. 
As one can see, except in STS-B where \DD-only slightly outperforms \modelshort, using \RR-only performs better than \DD-only. More importantly, when \modelrank~combines them it allows our \modelshort~to leverage both rankings and demonstrates improvement across all tasks. Note that the results of \DD-only also confirm that using only the output of the activation functions is not always sufficient for pruning and highlights the impact of using the trained model weights (More results in Figure~\ref{fig:-magnitude}).

\begin{figure*}[h!]
\centering
  \includegraphics[width=1.0\linewidth]{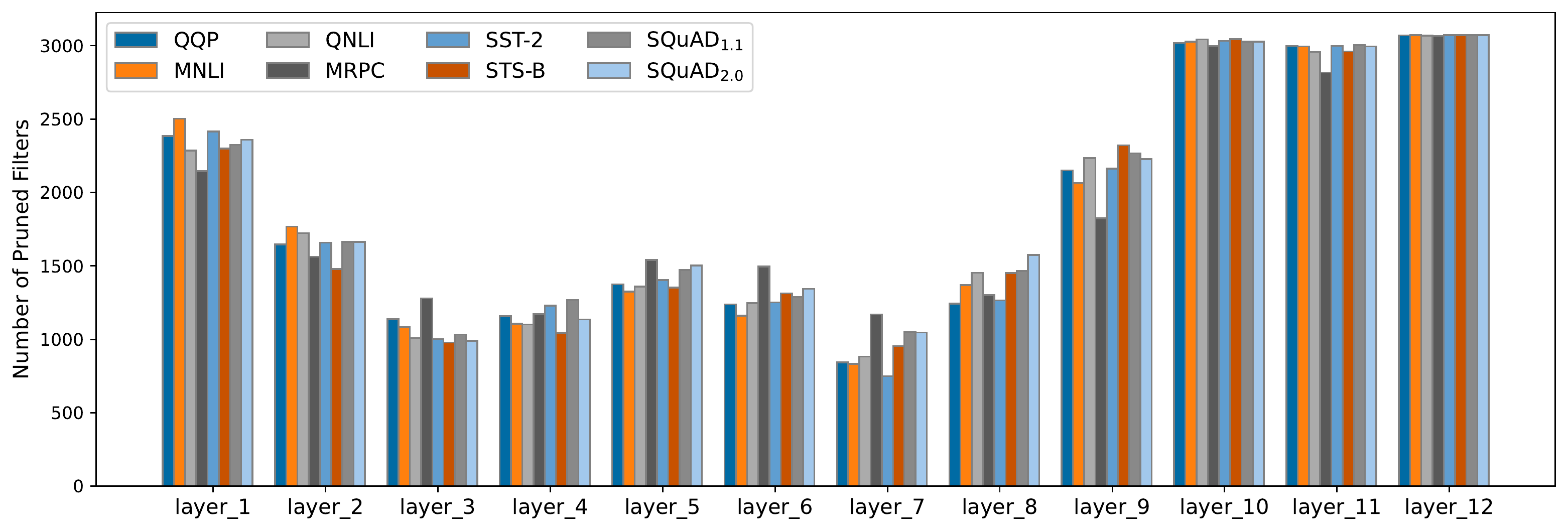}
  \caption{Ablation Study to investigate how many filters from each layer are pruned by \modelshort~on BERT$_{BASE}$. While middle layers seems to less get affected by KCM, many filters from the last three layers and first layer are pruned.}
  \label{fig:num_pruned}
\end{figure*}

\noindent{\bf Dynamic neuron selection}:
Another important feature of our \modelshort~is the fact that it dynamically decides how many neurons from each layer to prune. This feature is an outcome of merging the result of \modelrank~across all $L$ layers.
Figure~\ref{fig:num_pruned} illustrates how \modelshort~affects different layers of the BERT$_{BASE}$. Clearly more pruning occurs over the last three layers, and more than half of the filters in the first two layers are pruned. From \modelshort~point of view, the middle layers seem to be more important across all tasks.

\subsection{Discussion}
Our \modelshort~is a gradient-free structured pruning framework that neither requires retraining nor labeled data. Here we would like to discuss what if we have a limited labeled data and how our approach can be extended to leverage that. 

Recall that \modelrank~uses the statistics from the unlabeled data to rank filters based on layer-wise output. Thus a simple add-on would be to freeze the trained model, use the limited labeled data and only do one forward-backward pass and gather the gradient over the mask variables. Note that unlike Mask-Tuning~\cite{kwon2022fast}, we do not calculate the Fisher information since we just want to use the gradient as the new signal for the pruning. 
To do so, for example 1) the gradient information can be used as a new ranking criteria that can be combined into our \modelrank~or 2) one can use it to refine the top-k results of our \modelshort. Specifically, let us assume $f_i$ be the least important filter in top-k result of the \modelshort, and $f_j$ be the most important one from the gradient scores. If $f_j$ is not already in the top-k results, we can switch $f_i$ with $f_j$ if the total gradient of the top-k results increases. 
We implemented this simple greedy solution as an add-on to our \modelshort~and show that indeed having a limited labeled data contributes to improve the accuracy drop. Table~\ref{tab:distill-bert-hybrid} shows the result on DistilBERT where only 512 sampled label data is available. Since this is out of the scope of this work, We leave a more thorough investigation as a future work.  

\label{lbl:app-ext}
\begin{table*}[h!]
    \centering
    \caption{How few labeled data improves accuracy of our pruning framework \modelshort~on DistilBERT.}
    \label{tab:distill-bert-hybrid}
\resizebox{1.0\textwidth}{!}{%
    \begin{tabular}{c|c|cc|cc|cc|cc}
    \toprule
    \multirow{2}{*}{$!\nabla$} & \multirow{2}{*}{Method} &  \multicolumn{2}{c|}{QQP} & \multicolumn{2}{c|}{MNLI} & \multicolumn{2}{c|}{MRPC} & \multicolumn{2}{c}{QNLI} \\
    &  & 60\% & 70\%  & 60\% & 70\% & 60\% & 70\% & 60\%  & 70\% \\
    \hline
    & baseline & \multicolumn{2}{c|} {89.99} & \multicolumn{2}{c|} {82.11} & \multicolumn{2}{c|} {84.80} & \multicolumn{2}{c} {88.56} \\
    \hline
    \xmark & Mask-Tuning & $88.71 \pm 0.22$ & $89.66 \pm 0.06$ & $80.51 \pm 0.19$ & $81.65 \pm 0.09$ & $84.73 \pm 0.71$ & $84.83 \pm 0.35$ & $87.72 \pm 0.38$ & $88.43 \pm 0.07$  \\
    \hline
    \color{my-green} \checkmark & \bf \modelshort & $88.16 \pm 0.03$ & $89.28 \pm 0.03$ & $78.05 \pm 0.08$ & $80.60 \pm 0.05$ & $79.66 \pm 0.27$ & $83.01 \pm 0.16$ & $85.93 \pm 0.09$ & $86.93 \pm 0.13$ \\
    \hline
    \xmark & \bf Extension(512 labeled data) & $88.76 \pm 0.25$ & $89.45 \pm 0.07$ & $80.02 \pm 0.25$ & $81.37 \pm 0.11$ & $83.70 \pm 1.40$ & $84.49 \pm 0.49$ & $87.21 \pm 0.54$ & $88.21 \pm 0.15$  \\
    \xmark & \bf Extension(1k labeled data) & $88.92 \pm 0.20$ & $89.53 \pm 0.08$ & $80.41 \pm 0.11$ & $81.50 \pm 0.12$ & $84.17 \pm 0.45$ & $84.68 \pm 0.49$ & $87.60 \pm 0.31$ & $88.29 \pm 0.16$  \\
    \hline
    \hline
    \multirow{2}{*}{$!\nabla$} &\multirow{2}{*}{Method} &  \multicolumn{2}{c|}{SST-2} & \multicolumn{2}{c|}{STS-B} & \multicolumn{2}{c|}{SQuAD$_{1.1}$} & \multicolumn{2}{c}{SQuad$_{2.1}$} \\
    & &  60\% & 70\%  & 60\% & 70\% & 60\% & 70\% & 60\%  & 70\% \\
    \hline
    & baseline & \multicolumn{2}{c|} {91.40} & \multicolumn{2}{c|} {86.12}  & \multicolumn{2}{c|} {85.73} & \multicolumn{2}{c} {68.84}\\
    \hline
    \xmark & Mask-Tuning & $90.44 \pm 0.41$ & $90.93 \pm 0.24$ & $85.73 \pm 0.07$ & $85.96 \pm 0.10$ & $83.20 \pm 0.16$ & $84.64 \pm 0.09$ & $62.36 \pm 1.40$ & $65.32 \pm 0.48$\\
    \hline
    \color{my-green} \checkmark & \bf \modelshort & $88.38 \pm 0.25$ & $90.61 \pm 0.25$ & $85.26 \pm 0.02$ & $85.55 \pm 0.03$ & $76.92 \pm 0.11$ & $82.65 \pm 0.06$ & $64.56 \pm 0.11$ & $68.19 \pm 0.06$\\
    \hline
    \xmark & \bf Extension(512 labeled data) & $89.32 \pm 0.54$ & $90.38 \pm 0.35$ & $85.83 \pm 0.11$ & $86.02 \pm 0.07$ & $81.41 \pm 0.26$ & $83.30 \pm 0.09$ & $66.51 \pm 0.24$ & $67.72 \pm 0.18$  \\
    \xmark & \bf Extension(1k labeled data) & $89.86 \pm 0.56$ & $90.62 \pm 0.40$ & $85.90 \pm 0.11$ & $86.04 \pm 0.06$ & $81.16 \pm 0.22$ & $83.34 \pm 0.11$ & $66.35 \pm 0.32$ & $67.74 \pm 0.18$  \\
    \bottomrule
    \end{tabular}%
}
\end{table*}

\section{Related Work}\label{sec:rel_work}
There has been a lot of work on efficient transformers that improve inference speed and reduce memory usage, including efficient architecture design~\cite{kitaev2020reformer, iandola2020squeezebert, sun2020mobilebert, wang2020linformer, wu2020lite, fakoor2020trade, lan2019albert, lee2022littlebird}, neural architecture search~\cite{so2021searching, chen2020adabert, so2019evolved, wang2020hat, xu2021bert, yin2021autotinybert}, knowledge distillation~\cite{sun2020mobilebert, jiao2019tinybert, sanh2019distilbert, sun2019patient, fakoor2020fast}, quantization~\cite{kim2021bert, shen2020q, zadeh2020gobo, zafrir2019q8bert}, and hardware-software co-design~\cite{ham2021elsa, tambe2021edgebert, wang2021spatten, gu2022heat, shi2018neuroinspired}.

Pruning is an important area of research for model sparsity that removes insignificant weights in neural networks. While ~\citet{kurtic2022optimal, sanh2020movement, gale2019state, zhang2022platon} proposed second-order, first-order, and magnitude-based pruning methods for Transformers, ~\citet{chen2020lottery, chen2020earlybert, prasanna2020bert} explored the lottery ticket hypothesis. These methods can significantly reduce the model size; however, they might not offer significant inference speedup since the hardware and cannot efficiently utilize the unstructured sparse patterns.


Structured pruning methods, on the other hand, target removing groups of parameters. For example, low-rank factorization~\cite{gu2022heat, wang2019structured}, corsets based techniques~\cite{mussay2021data, mussay2019data, liebenwein2019provable, baykal2018data}, block-wise sparsity~\cite{li2020efficient}, and tile-wise sparsity~\cite{guo2020accelerating} prune structured sets of parameters in weight matrices. 
Additionally, attention head pruning~\cite{michel2019sixteen, voita2019analyzing} and layer dropping~\cite{fan2019reducing, sajjad2023effect, peer2022greedy} have been commonly used as more coarse-grained methods. Recent research has also explored combining different pruning granularity and principles to maximize model efficiency in all dimensions~\cite{chen2021chasing, khetan2020schubert, lagunas2021block, lin2020pruning, liu2021rosita, xia2022structured, yao2021mlpruning}. Another approach is to dynamically prune Transformers during inference time~\cite{fan2019reducing, hou2020dynabert, liu2021ebert, xin2020deebert, zhou2020bert}.

Even though structured pruning methods can be effective for compression and speedup, they can be difficult to implement in practice due to the high computational cost and complexity of the process. Additional training during or after pruning can be up to 10 times more expensive than original model training~\cite{lagunas2021block, xia2022structured}, and the pruning pipeline often requires rewriting the training code and involves many additional hyperparameters to adjust~\cite{hou2020dynabert, lan2019albert, liu2021rosita, yao2021mlpruning}. 

To tackle this, post-training model compression has been recently studied in~\citet{kwon2022fast, hubara2021accelerated, frantar2022optimal}. 
While~\citet{hubara2021accelerated, hubara2020improving, banner2019post} improves post training neural quantization,~\citet{kwon2022fast} proposed a fast post-training structured pruning framework for Transformers. Even though this approach avoids expensive retraining, it requires labeled data in the pruning pipeline.

Pruning in an unsupervised setting has been studied in~\citet{guo2020unsupervised, browne2020pulsenetone, browne2021unsupervised} for spiking neural networks and fully-connected layers; however, the pruning either happens during training or still requires retraining of the pruned model. 
In contrast, our structured pruning method neither requires retraining nor labeled data.

\section{Conclusion}
\label{sec:conclusion}
In this work, we studied the problem of structured pruning with unlabeled data and no backward pass.
We proposed a gradient-free structured pruning framework that prunes the filters with the help of our proposed \modelrank~that combines two ranking techniques called \emph{\RRfull} (\RR) and \emph{\DDfull} (\DD). We empirically evaluated our framework on GLUE and SQuAD benchmarks using BERT$_{BASE}$ and DistilBERT.  Compared to when the labeled data is available, our approach achieved up to 40\% FLOPs reduction with less than $4\%$ accuracy loss over all tasks considered. 

\nocite{langley00}

\bibliography{6-reference}
\bibliographystyle{icml2023}

\newpage
\appendix
\onecolumn
\section{Experimental Details}
\label{lbl:exp-detail-appendix}
We implemented our framework with PyTorch~\cite{pytorch} using the HuggingFace Transformers~\cite{hug_transformer} library.
We fine-tuned the pre-trained checkpoints of the BERT$_{BASE}$~\cite{devlin2018bert} and DistilBERT~\cite{sanh2019distilbert} downloaded from the HuggingFace repository on GLUE~\cite{wang2018glue} and SQuAD~\cite{rajpurkar2018know, rajpurkar2016squad} benchmarks.

GLUE~\cite{wang2018glue} includes following tasks.
1)Sentence similarity (QQP~\cite{shankar2017first}, MRPC~\cite{dolan2005automatically}, STS-B~\cite{cer2017semantic}) with 364K, 4k and 6k  training examples.
2) Sentiment classification (SST-2~\cite{socher2013recursive}) with 67K training example, 
3)Textual entailment (RTE~\cite{dagan2006pascal}) with 3K training examples. 
4) Natural language inference (MNLI~\cite{williams2017broad},
QNLI~\cite{rajpurkar2016squad}) with 392K, 105K training examples. 
We exclude CoLA~\cite{warstadt2019neural} and WLNI~\cite{levesque2012winograd} due to their unstable behaviors.
SQuAD 1.1~\cite{rajpurkar2016squad} and SQuAD 2.0~\cite{rajpurkar2018know} are question and answering tasks, each of which contains 88K and 130K training examples. SQuAD$_{2.0}$ is an extension of SQuAD$_{1.1}$ by including unanswerable questions whose answers are not stated in the given contexts.

All results are the averaged over the runs with 10 different seeds. For SQuAD tasks we report F1 score and for  all GLUE tasks except STS-B we report accuracy. For STS-B we report Spearman Correlation.

For Pruning we only use 2K raw data from the training sets. Note that we only use the input features and not their labels.
For our \RRfull~calculation, in Algorithm~\ref{alg:frr}, the width of Gaussian kernel $\sigma$, and convergence rate $\alpha$ are the hyperparameters. In our experiments, we set $\sigma=1.0$ and $\alpha=0.01$. On the average it only takes less than 20 iterations to converge.

\section{More Experimental Results}
\label{lbl:exp-results-appendix}

\subsection{Comparison with Gradient-free, Retrain-free, Supervision-free Baselines}
\begin{figure*}[h!]
\centering
  \includegraphics[width=0.95\linewidth]{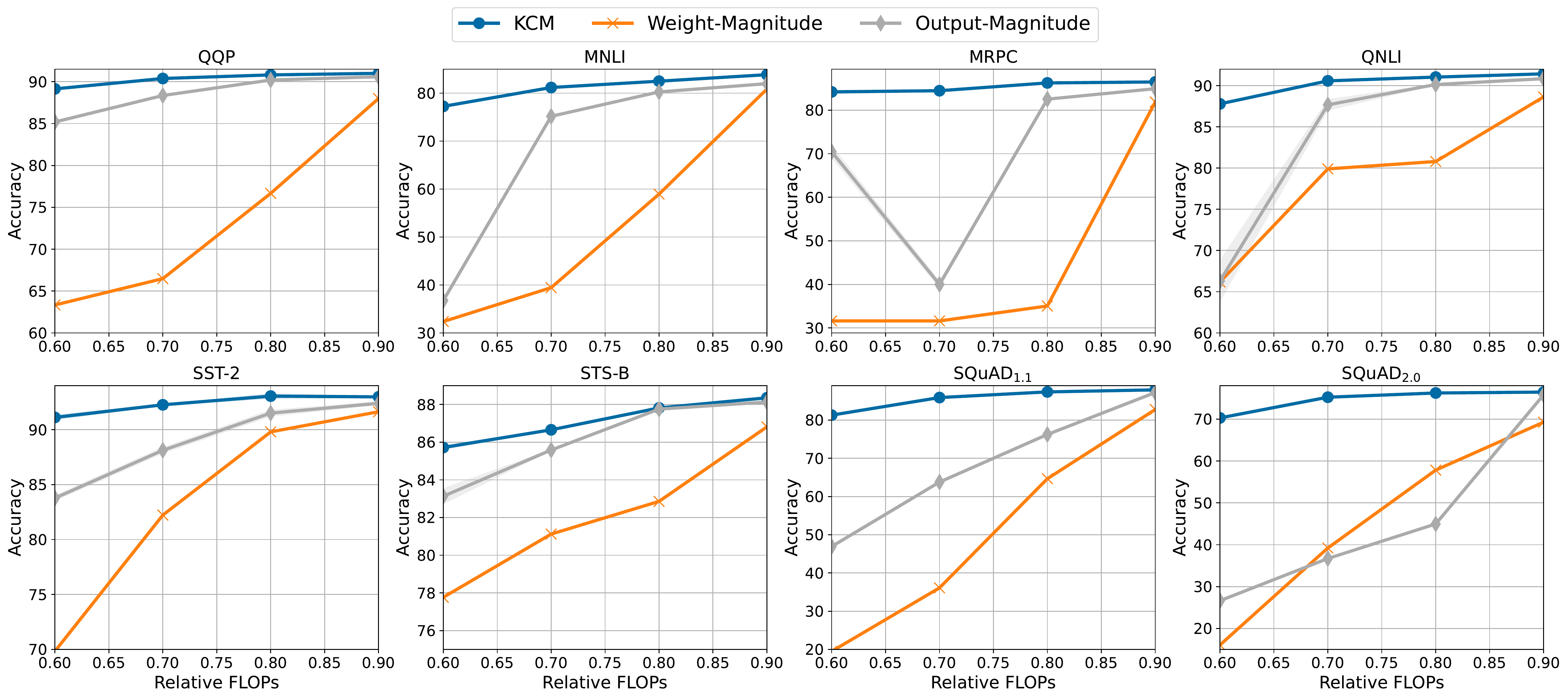}
  \vspace{-0.2in}
  \caption{\modelshort~outperforms Weight-magnitude, and outputs-magnitude across all GLUE and SQuAD tasks.}
  \label{fig:-magnitude}
\end{figure*}

Figure~\ref{fig:-magnitude} shows the performance of~\modelshort~against Weight-magnitude, and layer-wise Output-magnitude using BERT$_{BASE}$ on GLUE and SQuAD tasks. All these methods are gradient-free (no backward pass), retrain-free (no retrain), supervision-free (no labeled data), and run in a matter of minutes. Clearly, \modelshort~ outperforms across all tasks considered. 

\subsection{Impact of Unlabeled Data Sample Size}
\begin{figure*}[h!]
\centering
  \includegraphics[width=0.95\linewidth]{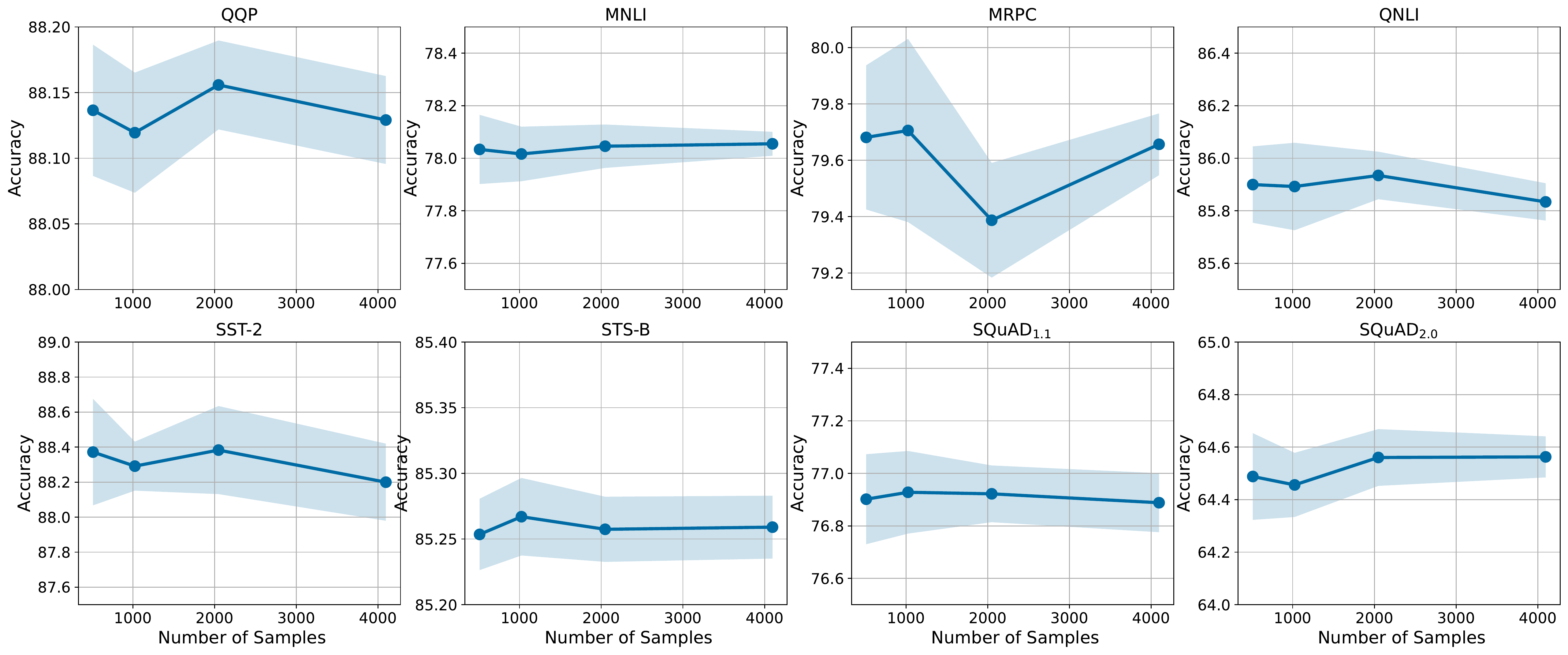}
  \vspace{-0.2in}
  \caption{Performance of \modelshort~on DistilBERT by varying unlabeled data sample size.}
  \label{fig:sample_num}
\end{figure*}

Figure~\ref{fig:sample_num} shows how the performance of \modelshort~on DistilBERT changes while we varied unlabeled data sample size.

\subsection{Speedup}
We evaluated the latency on real hardware and obtained the speedup of \modelshort~on BERT$_{BASE}$ on a single NVIDIA V100 GPU for 60\% Flops constraint:
\begin{table*}[h!]
    \centering
    \caption{Speedup of \modelshort~on BERT$_{BASE}$ on a single NVIDIA V100 GPU for 60\% Flops constraint.}
    \label{tab:bertlarge-extension}
\resizebox{0.75\textwidth}{!}{%
    \begin{tabular}{ccccccccc}
    \toprule
    Method & QQP & MNLI &  MRPC   & QNLI  &  SST-2 &  STS-B &  SQuAD$_{1.1}$ &  SQuAD$_{2.0}$\\
    \hline
    speedup & 1.58x &	1.46x	&  1.53x &   1.57x   &  1.58x &  1.59x  &   1.47x	 &  1.44x\\
    \bottomrule
    \end{tabular}%
}
\end{table*}

\section{SQuAD$_{1.1}$ Task on BERT$_{LARGE}$}
We conducted additional experiments on a larger-scale model, BERT$_{LARGE}$ over SQuAD$_{1.1}$. The results in Table~\ref{tab:bertlarge} indicate that our method outperforms unsupervised baselines, providing further evidence of its efficacy. 

\begin{table*}[h!]
    \centering
    \caption{\modelshort~outperforms Weight-magnitude, and Weight-magnitude-Scale on BERT$_{LARGE}$ over SQuAD$_{1.1}$.}
    \label{tab:bertlarge}
\resizebox{0.9\textwidth}{!}{%
    \begin{tabular}{ccccc}
    \toprule
    Method & 60\% & 70\% & 80\% & 90\%\\
    \hline
    Weight-Magnitude~\cite{li2016pruning} & 2.3029 & 2.8365 & 4.5763 & 52.5013\\
    Weight-Magnitude-Scale  & 2.3178 & 24.6811 & 83.6422 & 91.4438\\
    \bf \modelshort~(ours) & $\bf 85.86 \pm 0.14$	& $\bf 88.57\pm0.06$ &	$\bf 91.40 \pm 0.04$ & $\bf 92.72 \pm 0.03$\\
    \bottomrule
    \end{tabular}%
}
\end{table*}
\begin{table*}[h!]
    \centering
    \caption{How few labeled data improves accuracy of our pruning framework \modelshort~on BERT$_{LARGE}$ for SQuAD$_{1.1}$.}
    \label{tab:bertlarge-extension}
\resizebox{0.85\textwidth}{!}{%
    \begin{tabular}{ccccc}
    \toprule
    Method & 60\% & 70\% & 80\% & 90\%\\
    \hline
    \modelshort & $85.86\pm0.14$ &	$88.57\pm0.06$ &	$91.40\pm0.04$ &	$92.72\pm0.03$\\
    \bf Extension(512 labeled data)  & $89.44\pm0.1$ &	$91.85\pm0.06$ &	$92.40\pm0.05$ &	$93.00\pm0.01$\\
    \bf Extension(1k labeled data) & $89.48\pm0.16$ &	$91.92\pm0.07$ &	$92.69\pm0.02$ &	$93.17\pm0.04$\\
    \bottomrule
    \end{tabular}%
}
\end{table*}

\section{Train-Test Data Discrepancy}
We ran new experiments with a new dataset called \texttt{new-Wiki} to further evaluate the effectiveness of the proposed method under training-test data discrepancy. As outlined in~\citet{miller2020effect}, \texttt{new-Wiki} is different from the original SQuAD$_{1.1}$ dataset and was generated using the Wikipedia dataset.

We explored various scenarios involving the sampling of unlabeled data from datasets that differ from the evaluation dataset. In particular, we sample unlabeled data from 1) \texttt{SQuAD$_{1.1}$-train} 2) \texttt{SQuAD$_{1.1}$-val} or 3) \texttt{new-Wiki} and evaluate on \texttt{SQuAD$_{1.1}$-val}. As evident from the results in Table \ref{tab:Discrepancy}, sampling unlabeled data from \texttt{new-Wiki} (and evaluating on \texttt{SQuAD$_{1.1}$-val}) yielded improved performance compared to sampling from \texttt{SQuAD$_{1.1}$-train} and evaluating on \texttt{SQuAD$_{1.1}$-val}. This finding further supports our assertion regarding the applicability and effectiveness of our approach.
\begin{table*}[h!]
    \centering
    \caption{Train-Test data discrepancy}
    \label{tab:Discrepancy}
\resizebox{0.7\textwidth}{!}{%
    \begin{tabular}{cccc}
    \toprule
    Unlabeled Sample & Evaluation & 60\% & 70\%\\
    \hline
    \texttt{SQuAD$_{1.1}$-train}	& \texttt{SQuAD$_{1.1}$-val}	& $76.92 \pm 0.11$ &	$82.65 \pm 0.06$\\
    \texttt{SQuAD$_{1.1}$-val} &	\texttt{SQuAD$_{1.1}$-val} &	$77.17\pm0.073$	& $82.75\pm0.074$ \\
    \texttt{new-Wiki} &	\texttt{SQuAD$_{1.1}$-val} &	$77.45\pm0.076$ &	$82.80\pm0.031$\\
    \bottomrule
    \end{tabular}%
}
\end{table*}

We further evaluate the effectiveness of our approach on using the finetuned model on SQuAD$_{1.1}$ but evaluate on \texttt{new-Wiki} and results are as follows:
\begin{table*}[h!]
    \centering
    \caption{Performance of Finetuned model on SQuAD$_{1.1}$ on \texttt{new-Wiki} dataset.}
    \label{tab:Discrepancy2}
\resizebox{0.7\textwidth}{!}{%
    \begin{tabular}{cccc}
    \toprule
    Unlabeled Sample & Evaluation & 60\% & 70\%\\
    \hline
    \texttt{SQuAD$_{1.1}$-train}	& \texttt{new-Wiki} &	$75.26 \pm 0.08$ &	$81.14 \pm 0.096$\\
    \texttt{new-Wiki} &	\texttt{new-Wiki} &	$75.72\pm0.09$ &	$81.08\pm0.13$ \\
    \bottomrule
    \end{tabular}%
}
\end{table*}


\end{document}